\documentclass[journal]{IEEEtran}
\usepackage[utf8]{inputenc}
\usepackage{textcomp}
\usepackage{amssymb}
\usepackage{multirow} 
\usepackage{booktabs}
\usepackage{multirow}
\ifCLASSINFOpdf
\usepackage[table]{xcolor}
\definecolor{GoodGreen}{RGB}{198,226,188}
\definecolor{BadRed}{RGB}{222,159,165}

\usepackage{amsmath}
\usepackage{graphicx} 
\usepackage{subfigure}

\begin{document}
%
\title{Initial Model Incorporation for Deep Learning FWI: Pretraining or Denormalization?}
%
%
%

\author{Ruihua~Chen,  Bangyu~Wu,~\IEEEmembership{Member,~IEEE}, Meng~Li, Kai~Yang
\thanks{}
\thanks{Ruihua Chen, Bangyu Wu are with the School of Mathematics and Statistics, Xi’an Jiaotong University, Xi’an \emph{710049}, China (e-mail: Ruihua.chen@stu.xjtu.edu.cn; bangyuwu@xjtu.edu.cn).

Meng Li is with the Department of Oil and Gas Geophysics, CNPC Research Institute of Petroleum Exploration \& Development, Beijing \emph{100083}, China (e-mail: lmeng0203@petrochina.com.cn).

Kai Yang is with the School of Marine and Earth Science, Tongji University, Shanghai \emph{200092}, China (e-mail: yang\_kai@tongji.edu.cn)}
\thanks{}}

%
%

\markboth{IEEE Geoscience and Remote Sensing Letters}%
{Shell \MakeLowercase{\textit{et al.}}: Bare Demo of IEEEtran.cls for IEEE Journals}
%



\maketitle
\begin{abstract}
Subsurface property neural network reparameterized full waveform inversion (FWI) has emerged as an effective unsupervised learning framework, which can invert stably with an inaccurate starting model. It updates the trainable neural network parameters instead of fine-tuning on the subsurface model directly. There are primarily two ways to embed the prior knowledge of the initial model into neural networks, that is, pretraining and denormalization. Pretraining first regulates the neural networks' parameters by fitting the initial velocity model; Denormalization directly adds the outputs of the network into the initial models without pretraining. In this letter, we systematically investigate the influence of the two ways of initial model incorporation for the neural network reparameterized FWI. We demonstrate that pretraining requires inverting the model perturbation based on a constant velocity value (mean) with a two-stage implementation. It leads to a complex workflow and inconsistency of objective functions in the two-stage process, causing the network parameters to become inactive and lose plasticity. Experimental results demonstrate that denormalization can simplify workflows, accelerate convergence, and enhance inversion accuracy compared with pretraining. 
\end{abstract}


\IEEEpeerreviewmaketitle

\begin{figure*}
    \centering
    \includegraphics[width=0.9\linewidth]{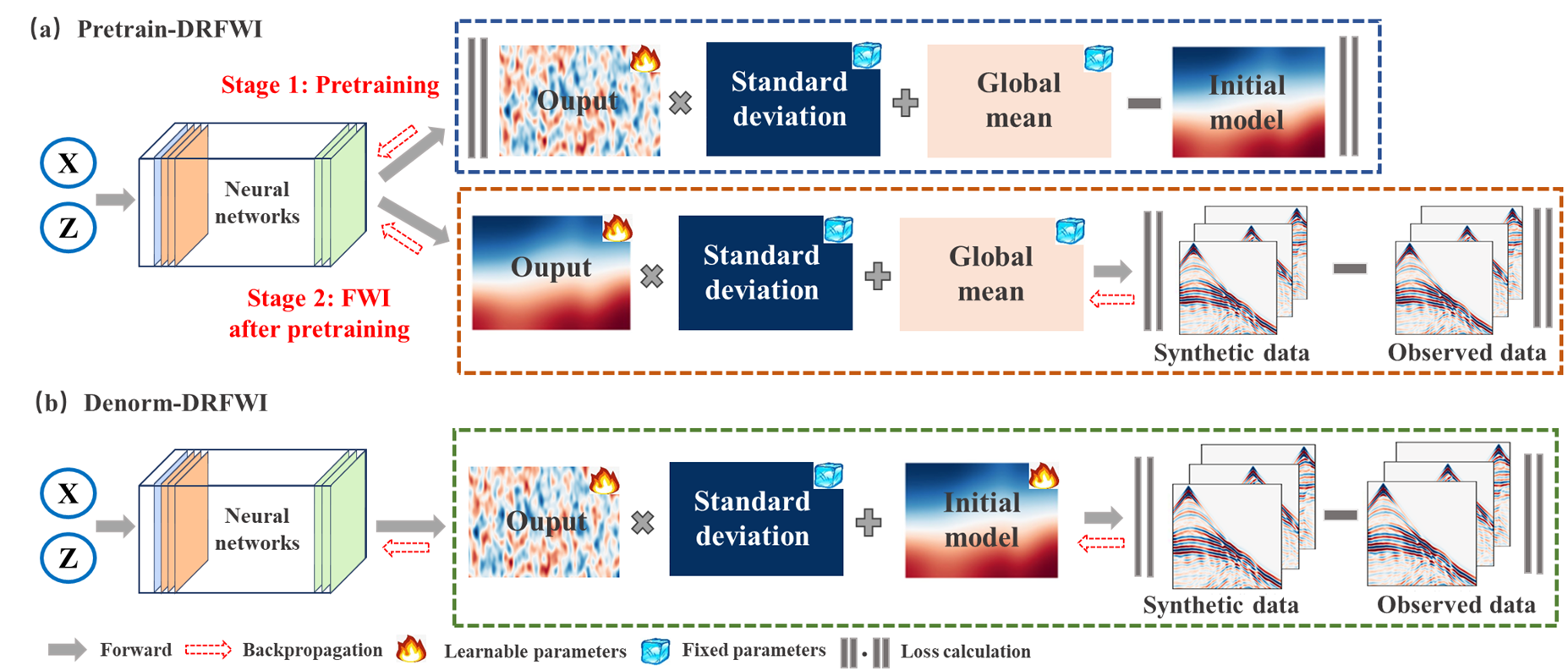}
    \caption{Initial Model Incorporation for DRFWI (e.g., Implicit FWI \cite{cite1}) with pretraining and denormalization. (a) Pretrain-DRFWI includes two stages (i.e., pretraining and FWI); (b) Denorm-DRFWI replaces the global mean matrix with the initial model to simplify the workflow.}
    \label{fig: workflow}
\end{figure*} 

\section{Introduction}

\IEEEPARstart{F}{ull} waveform inversion (FWI) is a high-precision subsurface imaging technique, which updates the parameter models using optimization algorithms to minimize the discrepancy between observed and synthetic seismic data \cite{cite2}. However, the complexity and non-linearity of the wave propagation mechanism, the non-convexity of the objective function, and limited seismic data lead to the ill-posedness of FWI. It necessitates an accurate initial model or regularization techniques to avoid the inversions into local minima \cite{cite3}. 

Deep learning reparameterized techniques have recently been used in FWI (termed DRFWI) to reduce the dependency on an accurate initial velocity model and enhance its robustness \cite{cite4}. DRFWI reparameterizes the velocity model with neural networks, mapping the specific input domain (e.g, wavefield via CNN-FWI \cite{cite5}, random noise via DIP-FWI \cite{cite6}, and coordinates via IFWI \cite{cite1}) to the velocity model domain, and integrates the PDE (e.g., acoustic wave equation) to embed the constraint of physics and reduce the dependency on the label data. 

When the initial velocity model (e.g., smooth or linear velocity models) is available, DRFWI integrates this prior knowledge through two distinct methods: (1) DRFWI with pretraining (\textbf{Pretrain-DRFWI})  \cite{cite1}. The neural network first undergoes supervised training to fit the initial velocity model, thereby encoding its structural features through parameter optimization. (2) DRFWI with denormalization (\textbf{Denorm-DRFWI}) \cite{cite6}. Instead of pretraining, this approach directly adds the networks' output to the initial velocity model instead of the uniform model (mean) during denormalization. The pretraining strategy transfers knowledge via parameter learning, whereas the denormalization framework imposes physical constraints through architectural design. A comprehensive analysis of the efficacy of these two ways is essential, particularly when utilizing inaccurate starting models.

In this letter, we systematically investigate the influence of the two ways of initial model incorporation for the DRFWI, using Implicit FWI (IFWI) \cite{cite1} as an example. We demonstrate that the pretraining method has three limitations that hinder the practical implementation: 
\begin{itemize}
    \item Pretrain-DRFWI involves a two-stage implementation, which requires first fitting the initial model and then fitting the observed seismic data. This multi-step workflow increases complexity \cite{cite6} and introduces sensitivity to pretraining hyperparameters. 
    \item Pretrain-DRFWI necessitates inversion of a complete velocity model, including the low-frequency initial model and higher-frequency perturbations. This inversion tends to slow convergence and compromises high-frequency representation, primarily due to neural networks' inherent low-frequency bias \cite{cite7}.
    \item The approach faces challenges from the domain discrepancy between velocity and seismic fitting. When combined with inaccurate initial models, these factors can lead to negative transfer effects \cite{cite8}, where knowledge from the initial model adversely impacts FWI performance.
\end{itemize}
In contrast, Denorm-DRFWI can simplify the workflow by replacing the mean matrix with the initial velocity model. This modification can eliminate the risk of negative transfer effects and mitigate the inherent low-frequency bias of neural networks. By preserving network plasticity, the Denorm-DRFWI enhances both computational efficiency and inversion accuracy, ultimately yielding superior results compared with Pretrain-DRFWI.

\section{Methodology}\label{sec: method}
Distinct from updating the velocity model directly, DRFWI generates the model using neural networks with learnable weight parameters. The two main ways to embed initial prior knowledge are Pretrain-DRFWI and Denorm-DRFWI. In this section, we introduce the theory of these two approaches.  

\subsection{DRFWI with pretraining}
As shown in Fig. \ref{fig: workflow} (a), Pretrain-DRFWI includes two stages. In the first stage, Pretrain-DRFWI regulates the parameters of neural networks by minimizing the discrepancy between the predicted velocity model and the initial velocity model. Hence, the mathematical formulation of optimization is as follows:
\begin{align}
\min_{\mathbf{\Theta}}\mathcal{L}_1 = \frac{1}{2}\Big|\Big|m_{init} - {\cal D}_1\big(f_{\Theta}(I)\big) \Big|\Big|^2_F, \label{equ_stage1}
\end{align}
where $m_{init}$ denotes the initial velocity model, $f_{\Theta}(\cdot)$ represents a neural network with parameters $\Theta$, which maps the input domain $I$ (e.g., wavefields via CNN-FWI, random noise via DIP-FWI, and coordinates via IFWI) into the normalized model $f_{\Theta}(I)$. Additionally, ${\cal D}_1(\cdot)$ is the denormalization operator, which transforms the normalized model into the predicted velocity model as follows:
\begin{align}
    {\cal D}_1\big(f_{\Theta}(I)\big) :=  f_{\Theta}(I)\odot S + M\label{equ:denormalization}
\end{align}
where $f_{\Theta}(I)$ represents the normalized model, $S$ denotes the standard variance matrix, $M$ is the constant velocity value (i.e., mean), and $\odot$ denotes the Hadamard product. In the second stage, Pretrain-DRFWI inverts the complete velocity model by minimizing the discrepancy between the observed and synthetic seismic data. Hence, the mathematical formulation of optimization is as follows:
\begin{align}
\min_{\mathbf{\Theta}}\mathcal{L}_2 = \frac{1}{2}\Big|\Big|d_{obs} - {\cal M}\odot \mathcal{F}\big({\cal D}_1[f_{\Theta}(I)]\big) \Big|\Big|^2_F,\label{equ_stage2}
\end{align}
where $d_{obs}$ is observed seismic data, ${\cal F}(\cdot)$ represents the forward operator formed by a recurrent neural network (RNN) or finite difference operators, which maps the predicted velocity model into synthetic full-wavefield data, ${\cal M}$ is a mask matrix of the acquisition system. 

\subsection{DRFWI with denormalization}

Another approach is Denorm-DRFWI. As illustrated in Fig. \ref{fig: workflow} (b), Denorm-DRFWI decouples the initial velocity model from the neural networks by replacing the mean matrix $M$ with the initial velocity model $m_{init}$. The denormalization operator ${\cal D}_1(\cdot)$ in equation \ref{equ:denormalization} can thus be expressed as: 
\begin{align}
    {\cal D}_{2}\big(f_{\Theta}(I),m_{init}\big) :=  f_{\Theta}(I)\odot S + m_{init}\label{equ:denormalization}
\end{align}
where $m_{init}$ denotes the initial velocity model. Depending on whether $m_{init}$ is treated as fixed or learnable, we divide the Denorm-DRFWI into two distinct variants: Static Denorm-DRFWI (\textbf{S-Denorm-DRFWI}) and Adaptive Denorm-DRFWI (\textbf{A-Denorm-DRFWI}), respectively. 

Denorm-DRFWI can decouple the initial model from the neural networks, allowing the networks to focus solely on representing the perturbed model based on the initial velocity model. This approach provides two key advantages: (1) By streamlining the workflow from a two-stage to a single-stage process, Denorm-DRFWI eliminates issues of negative transfer and mitigates plasticity loss, thereby enhancing inversion accuracy; (2) Denorm-DRFWI only needs to represent the high-frequency perturbed model on the initial velocity model. It can alleviate the low-frequency bias of neural networks, as high-frequency components contribute more significantly to gradient updates. This approach can accelerate convergence and enable more detailed characterization of velocity structures.

\begin{figure}
    \centering
    \includegraphics[width=0.9\linewidth]{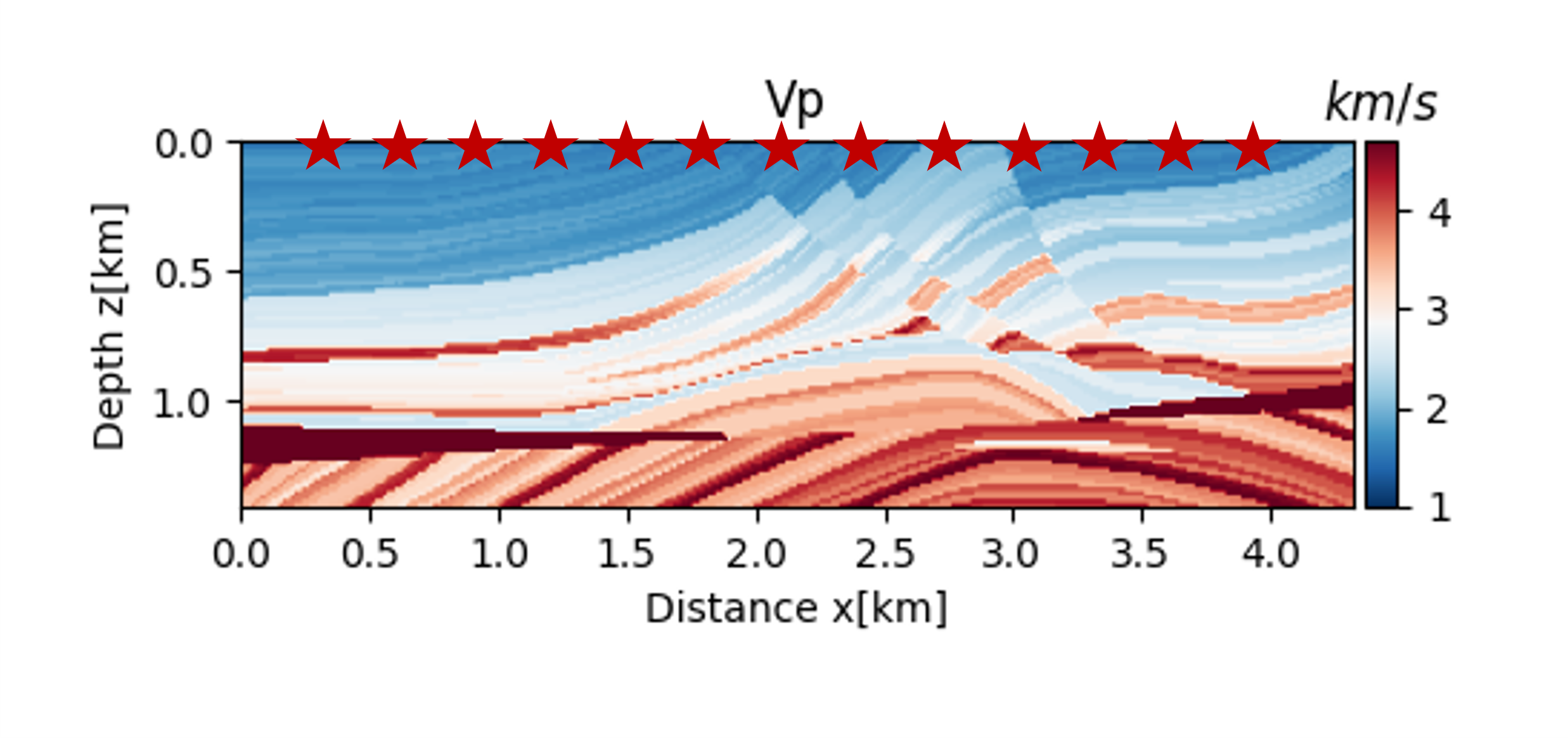}\vspace{-0.7cm}
    \caption{Marmousi model. The red stars denote the source locations.}
    \label{fig:marmousi}
\end{figure}

\section{Results}\label{sec:result}
This section investigates the inversion results of Pretrain-DRFWI and Denorm-DRFWI using coordinate-based inputs (i.e., IFWI \cite{cite1}). We first present our experimental configuration, including the dataset, forward modeling parameters, and neural network architecture. Then, we conduct a comprehensive comparison of inversion performance among Pretrain-DRFWI, S-Denorm-DRFWI, and A-Denorm-DRFWI approaches using both smooth and linear initial velocity models. Finally, we employ the low-frequency bias and negative transfer effects to provide mechanistic explanations.

\subsection{Experiment setting}
The experiments employ a 2D Marmousi model with a computational grid composed of 94 rows and 288 columns, featuring uniform horizontal and vertical grid spacing of 15 m. For forward modeling, a Ricker wavelet with a dominant frequency of 8 Hz serves as the seismic source. Wavefield propagation is discretized with a temporal sampling interval of 1.9 ms, generating 1,000 recorded time samples. The upper boundary is configured with a free-surface boundary condition, while the remaining boundaries utilize Perfectly Matched Layers (PML) to mitigate spurious reflections. A total of 13 sources are spaced at 300 m intervals along the 15 m depth horizon, as illustrated by the red stars in Fig. \ref{fig:marmousi}, with receivers deployed at every surface grid node with 15 m spacing. All numerical simulations and inversion procedures were executed under identical computational configurations, implemented on an RTX 3090 GPU using the PyTorch deep learning framework.

Following the source code of IFWI \cite{cite1}, we utilize the following hyperparameter settings for Pretrain-DRFWI and Denorm-DRFWI. The activation function of the MLP is set as the sine function, and $\omega$ is set as 30. The depth of the MLP is set as 4. The number of neurons of the MLP is set to 128. The Adam optimizer is used to optimize the MLP. The learning rate is set as 0.0001 for Pretrain-DRFWI and Denorm-DeRFWI. We set 1 $\rm km/s$ and 3 $\rm km/s$ for the standard deviation matrix and global mean matrix, respectively.
\begin{figure}
    \centering
    \includegraphics[width=1\linewidth]{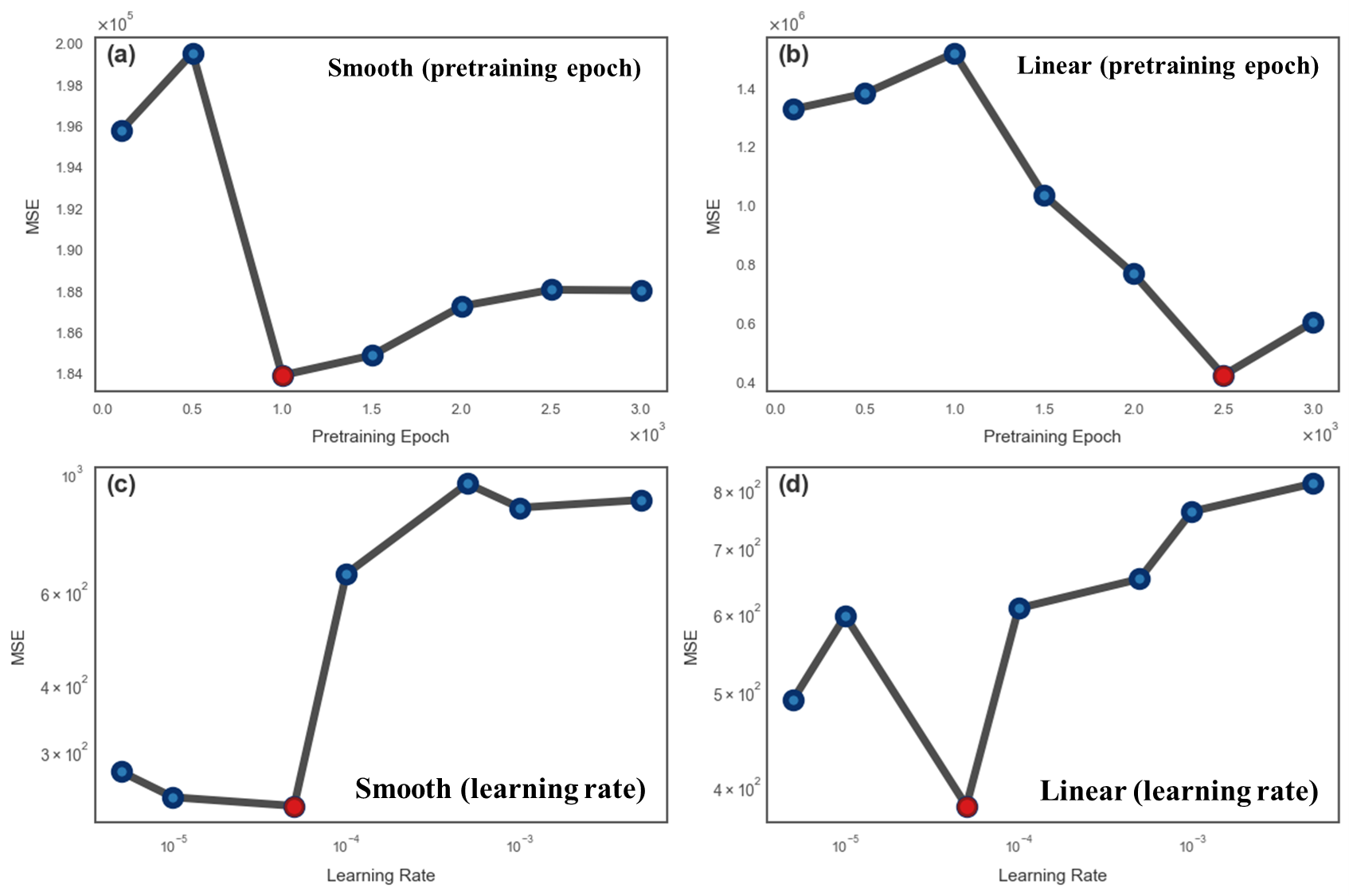}
    \caption{Pretraining hyperparameter analysis for Pretrain-DRFWI: (a) and (b) MSE vs. \textbf{pretraining epochs} (100 to 3000) with smooth and linear initial models, respectively (fixed lr = 5e-5); (c) and (d) MSE vs. \textbf{learning rate} (5e-6 to 5e-3) with smooth and linear initial models, respectively (using optimal epochs). Red points indicate optimal settings.}
    \label{fig:pretrainig_analysis}
\end{figure} 
\begin{figure*}
    \centering
    \includegraphics[width=1\linewidth]{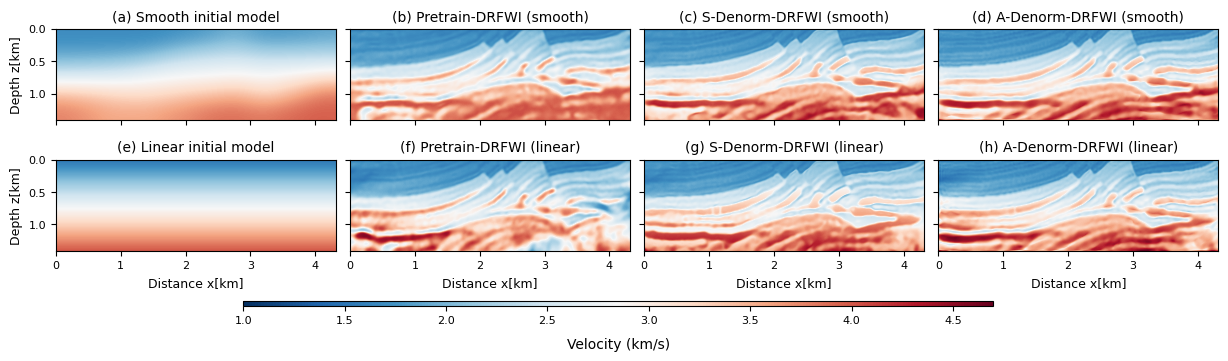}
    \caption{Comparisons of different methods using different initial models. (a) Smooth initial model; (e) Linear initial model; (b) - (d) Inversion results from Pretrain-DRFWI, S-Denorm-DRFWI, and A-Denorm-DRFWI using the smooth model, respectively; (f) - (h) Corresponding results using the linear model.}
    \label{fig:result_fwi}
\end{figure*}
\begin{figure}
    \centering
    \includegraphics[width=0.85\linewidth]{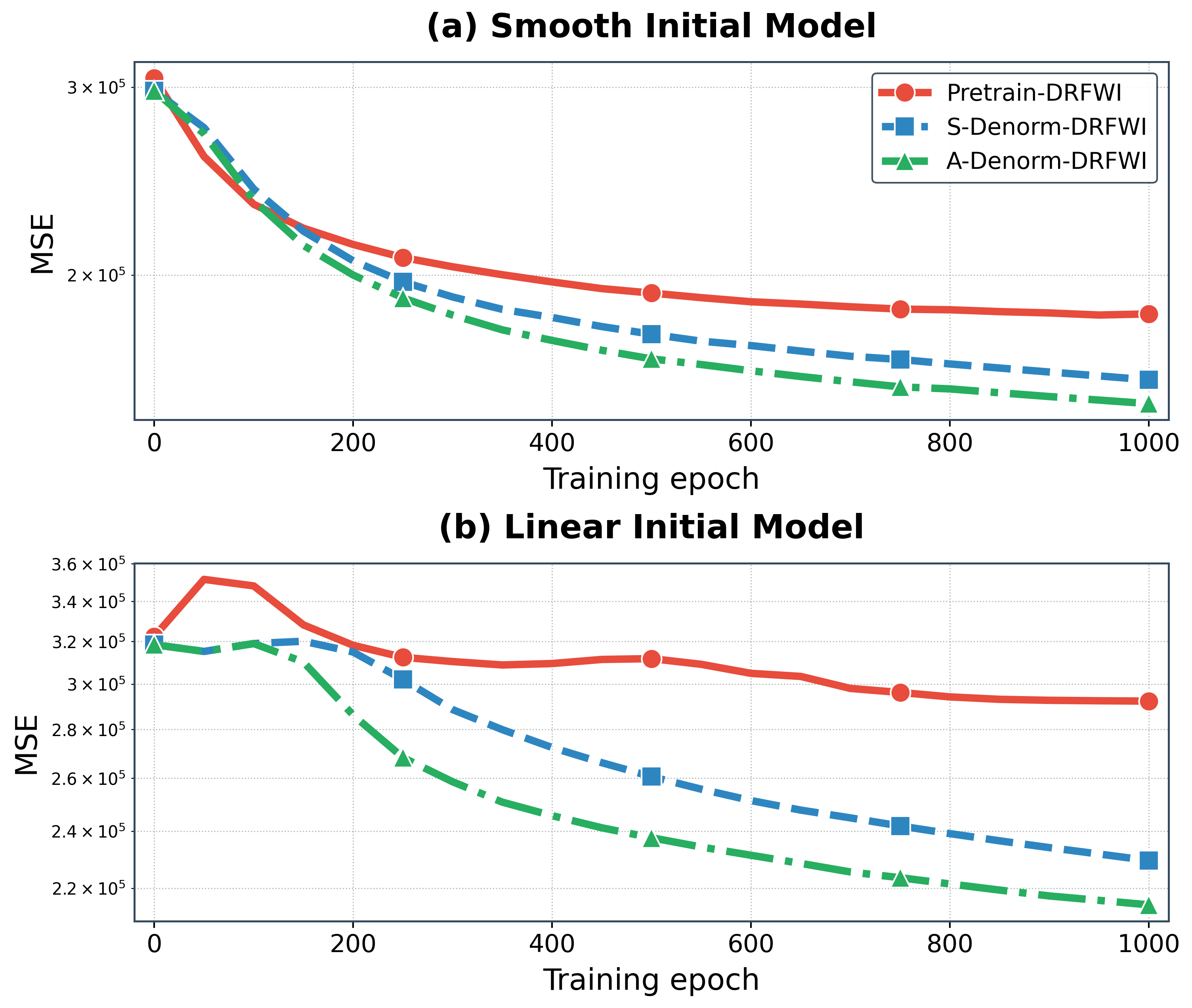}
    \caption{The error of model w.r.t. training epochs using different methods. (a) Error curve for smooth initial model; (b) Error curve for linear initial model.}
    \label{fig:loss_ifwi}
\end{figure}
\subsection{DRFWI with pretraining}

Since the original pretraining parameters for Pretrain-DRFWI \cite{cite1} are unavailable, we empirically optimize its hyperparameters (pretraining epochs and learning rate) by evaluating the MSE between the inverted model (after 1000 epochs) and ground truth, using both smooth and linear initial models.

Fig. \ref{fig:result_fwi} (a) shows the smooth initial model obtained by Gaussian smoothing of the ground truth, while (b) displays the linear initial model ranging from 1.5 $ \rm km/s$ to 4 $\rm km/s$. Fig. \ref{fig:pretrainig_analysis} (a) and (b) compare the MSE between inverted and ground-truth velocity models across different pretraining epochs (100-3000) for smooth and linear initial models, respectively. Results indicate optimal pretraining at 1000 epochs for the smooth model and 2500 epochs for the linear model (learning rate = 5e-5). Fig. \ref{fig:pretrainig_analysis} (c) and (d) show that 5e-5 yields the lowest MSE between inverted and true velocity models across both initial models at their optimal pretraining epochs. Thus, we use a pretraining learning rate of 5e-5, with 1000 epochs for the smooth initial model and 2500 for the linear initial model.

These results demonstrate that the final inversion results of Pretrain-DRFWI exhibit significant sensitivity to pretraining hyperparameters, necessitating extensive hyperparameter tuning in practical applications. We will show that the different objective functions in the two stages can cause negative transfer to the FWI stage and make the neural network parameters inactive and loss of plasticity in Sec. \ref{subsubsection: negative_transfer}. Moreover, Fig. \ref{fig:result_fwi} (b) and (f) denote the final inverted model using optimal pretraining hyperparameters with the smooth and linear initial velocity model, respectively. It is observed that the inversion results of Pretrain-DRFWI can not capture the details of deep velocity structure and are sensitive to the inaccurate initial model.

\subsection{DRFWI with denormalization} 

To evaluate the performance of Denorm-DRFWI, we conduct numerical experiments comparing Pretrain-DRFWI, S-Denorm-DRFWI, and A-Denorm-DRFWI using a smooth and linear initial velocity model with 1000 training epochs. Fig. \ref{fig:result_fwi} (b) - (d) represent the inversion results of different methods with the smooth initial model, respectively. We can observe that both S-Denorm-DRFWI and A-Denorm-DRFWI can recover more high-frequency details in the deep velocity structure compared with Pretrain-DRFWI. This can be attributed to the fact that the neural network of Denorm-DRFWI only needs to represent the high-frequency perturbed model based on the smooth initial model, thereby reducing the contribution of low-frequency gradients and mitigating the influence of the low-frequency bias. Fig. \ref{fig:result_fwi} (f) - (h) represent the inversion results of different methods with the linear initial model, respectively. We can observe that Denorm-DRFWI is more robust to inaccurate initial velocity models and can achieve better inversion results even under the inaccurate initial model.

Fig. \ref{fig:loss_ifwi} (a) and (b) illustrate the velocity model error curves for smooth and linear initial models w.r.t. training epochs, respectively. We can observe that Denorm-DRFWI achieves faster convergence and lower velocity model error compared with Pretrain-DRFWI, while A-Denorm-DRFWI obtains the most accurate inversion results. As shown in Table \ref{tab:comparison}, the comprehensive evaluation metrics of different methods confirm that A-Denorm-DRFWI outperforms all other methods across all metrics, demonstrating its stability and effectiveness.
\begin{table}
\centering
\small
\setlength{\tabcolsep}{5pt}
\caption{
    Performance comparison for different methods. \textbf{Bold} values indicate superior performance within the same initial model. ($\uparrow$: higher better, $\downarrow$: lower better)
    \label{tab:comparison}
}
\begin{tabular}{@{}llcccc@{}}
\toprule
\multirow{2}{*}{Model} & \multirow{2}{*}{Method} & \multicolumn{4}{c}{Evaluation metrics} \\ 
\cmidrule(lr){3-6}
& & \multicolumn{1}{c}{MSE$\downarrow$} & \multicolumn{1}{c}{MAE$\downarrow$} & \multicolumn{1}{c}{R$^2$$\uparrow$} & \multicolumn{1}{c}{SSIM$\uparrow$} \\ 
\midrule

\multirow{3}{*}{Smooth} 
 & Pretrain-DRFWI    & 0.1839 & 0.2371 & 0.6309 & 0.8016  \\
 & S-Denorm-DRFWI    & 0.1605 & 0.2206 & 0.6791 & 0.8268 \\
 & A-Denorm-DRFWI    & \textbf{0.1517} & \textbf{0.2146} & \textbf{0.6899} &\textbf{ 0.8363}  \\
\addlinespace[2pt]

\multirow{3}{*}{Linear}
 &Pretrain-DRFWI  & 0.3945 & 0.3604 & 0.5574 & 0.5745  \\
 & S-Denorm-DRFWI    & 0.2148 & 0.2636 & 0.6147 & 0.7683 \\
 & A-Denorm-DRFWI    & \textbf{0.2002} & \textbf{0.2518} & \textbf{0.6328} & \textbf{0.7816}  \\
 
\bottomrule
\end{tabular}
\end{table}

\subsection{Low-frequency bias and negative transfer}

Our analysis reveals three key limitations of Pretrain-DRFWI: (1) sensitivity to pretraining hyperparameters, (2) poor recovery of deep high-frequency structures, and (3) weaker robustness to inaccurate initial models versus Denorm-DRFWI. We explain these phenomena through low-frequency bias and negative transfer mechanisms.

\subsubsection{Low-frequency bias}\label{subsubsection: low-frequency}

\begin{figure}
    \centering
    \includegraphics[width=0.90\linewidth]{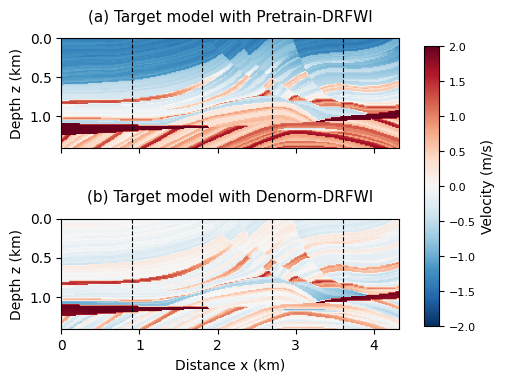}\vspace{-0.2cm}
    \caption{Target fitting model. (a) Pretrain-DRFWI; (b) Denorm-DRFWI}
    \label{fig:target_model}
\end{figure}

\begin{figure}
    \centering
    \includegraphics[width=0.95\linewidth]{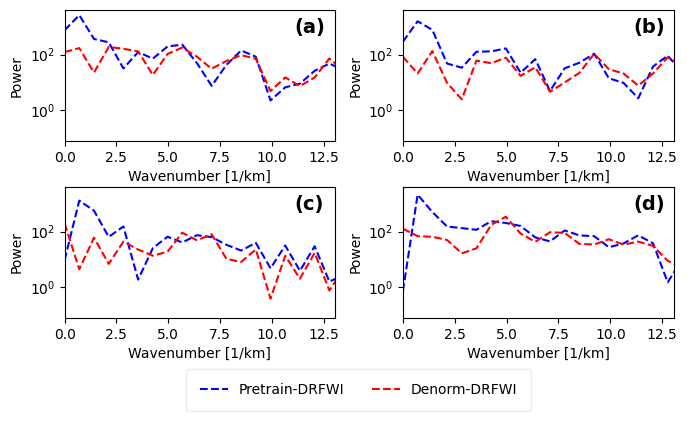}

    \caption{Wavenumber spectrum along four distinct velocity
profiles at different lateral locations using Pretrain-DRFWI and Denorm-DRFWI.}
    \label{fig:wavenumber}
\end{figure}

    Low-frequency bias means that in neural network training, low-frequency components dominate the learning due to their stronger and more stable gradients, while high-frequency components learn more slowly or unstably \cite{cite7}. Fig. \ref{fig:target_model} (a) - (b) represents the target fitting model using Pretrain-DRFWI and Denorm-DRFWI, respectively. We can observe that Pretrain-DRFWI needs to invert the complete model, including the low-frequency initial model, while Denorm-DRFWI only needs to invert the high-frequency perturbed model. Fig. \ref{fig:wavenumber} (a) - (d) denotes the wavenumber spectrum along four distinct velocity profiles at different lateral locations (marked by black dashed lines in Fig. \ref{fig:target_model}). We can observe that Denorm-DRFWI needs to invert fewer low-frequency components, exhibiting lower relative low-frequency gradients but higher relative high-frequency gradients compared with Pretrain-DRFWI, thereby achieving faster high-frequency learning rates. Hence, Denorm-DRFWI can invert more high-frequency details in the deep velocity structure with the smooth and linear initial model, as shown in Fig. \ref{fig:result_fwi}.

\subsubsection{Negative transfer} \label{subsubsection: negative_transfer}
Pretrain-DRFWI employs transfer learning, pretraining on the initial velocity model, then learning the perturbation model based on the initial model. However, we observe that knowledge transfer from the initial model can negatively impact the FWI stage in Figs. \ref{fig:pretrainig_analysis} and \ref{fig:result_fwi}. To investigate this, we compute the Euclidean Distance and Cosine Similarity between network parameters. The Cosine Similarity is defined as:
\begin{align}
CS(A,B) = \frac{A\cdot B}{||A||_2||B||_2}=\frac{\sum_{i=1}^{n}A_iB_i}{\sqrt{\sum_{i=1}^nA_i^2}\sqrt{\sum_{i=1}^nB_i^2}}
\end{align}
where $A, B\in \mathbb{R}^n$ are parameter vectors. Here, $CS(A,B) = 1$ indicates identical directions, while $CS(A,B)=-1$ denotes opposite directions.  Table \ref{tab:layer_params_en} presents the Distance and Cosine Similarity measurements for Pretrain- and Denorm-DRFWI.

The observations reveal that the first-stage parameter distribution of Pretrain-DRFWI aligns closely with the initial model, whereas Denorm-DRFWI exhibits an opposite trend, indicating divergent optimization paths due to distinct objective functions. Furthermore, the second-stage parameters of Pretrain-DRFWI show minimal adjustments compared with the first stage, suggesting that the initial training traps the model in a local optimum near the starting point (i.e., loss of plasticity \cite{cite9}), limiting its ability to capture high-frequency waveform details. Hence, the opposing optimization paths and loss of plasticity lead to negative transfer \cite{cite8}, making the results sensitive to pretraining hyperparameters and yielding poorer inversion details.

\begin{table}
\centering
\caption{Comparison of Cosine Similarity (CS) and Euclidean Distance (ED) for Pretrain-DRFWI (two-stage) and Denorm-DRFWI using smooth initial model.}
\label{tab:layer_params_en}
\footnotesize
\begin{tabular}{l *{6}{r}}
\toprule
\multirow{2}{*}{Parameters} & 
\multicolumn{4}{c}{Pretrain-DRFWI} & 
\multicolumn{2}{c}{Denorm-DRFWI} \\
\cmidrule(lr){2-3} \cmidrule(lr){4-5} \cmidrule(lr){6-7}
 & \multicolumn{2}{c}{Stage 1 vs INI} & \multicolumn{2}{c}{Stage 1 vs Stage 2} & \multicolumn{2}{c}{FWI vs INI} \\
\cmidrule(lr){2-3} \cmidrule(lr){4-5} \cmidrule(lr){6-7}
 & CS & ED & CS & ED & CS & ED \\
\midrule

L0.weight & 
\cellcolor{GoodGreen!100.00}1.000 & 
\cellcolor{BadRed!0.40}0.008 & 
\cellcolor{GoodGreen!100.00}1.000 & 
\cellcolor{BadRed!0.30}0.006 & 
\cellcolor{GoodGreen!45.45}-0.091 & 
\cellcolor{BadRed!76.05}1.521 \\

L0.bias & 
\cellcolor{GoodGreen!100.00}1.000 & 
\cellcolor{BadRed!0.25}0.005 & 
\cellcolor{GoodGreen!100.00}1.000 & 
\cellcolor{BadRed!0.15}0.003 & 
\cellcolor{GoodGreen!48.50}-0.030 & 
\cellcolor{BadRed!71.00}1.420 \\

L1.weight & 
\cellcolor{GoodGreen!50.10}0.002 & 
\cellcolor{BadRed!21.90}0.438 & 
\cellcolor{GoodGreen!52.05}0.041 & 
\cellcolor{BadRed!19.00}0.380 & 
\cellcolor{GoodGreen!49.50}-0.010 & 
\cellcolor{BadRed!81.70}1.634 \\

L1.bias & 
\cellcolor{GoodGreen!100.00}1.000 & 
\cellcolor{BadRed!1.35}0.027 & 
\cellcolor{GoodGreen!100.00}1.000 & 
\cellcolor{BadRed!0.70}0.014 & 
\cellcolor{GoodGreen!48.50}-0.030 & 
\cellcolor{BadRed!72.35}1.447 \\

L2.weight & 
\cellcolor{GoodGreen!96.75}0.935 & 
\cellcolor{BadRed!18.05}0.361 & 
\cellcolor{GoodGreen!99.00}0.980 & 
\cellcolor{BadRed!10.80}0.216 & 
\cellcolor{GoodGreen!49.85}-0.003 & 
\cellcolor{BadRed!78.05}1.561 \\

L2.bias & 
\cellcolor{GoodGreen!100.00}1.000 & 
\cellcolor{BadRed!0.90}0.018 & 
\cellcolor{GoodGreen!100.00}1.000 & 
\cellcolor{BadRed!0.30}0.006 & 
\cellcolor{GoodGreen!40.00}-0.200 & 
\cellcolor{BadRed!77.75}1.555 \\

L3.weight & 
\cellcolor{GoodGreen!96.65}0.933 & 
\cellcolor{BadRed!18.55}0.371 & 
\cellcolor{GoodGreen!99.65}0.993 & 
\cellcolor{BadRed!6.35}0.127 & 
\cellcolor{GoodGreen!50.20}0.004 & 
\cellcolor{BadRed!76.50}1.530 \\

L3.bias & 
\cellcolor{GoodGreen!100.00}1.000 & 
\cellcolor{BadRed!0.90}0.018 & 
\cellcolor{GoodGreen!100.00}1.000 & 
\cellcolor{BadRed!0.15}0.003 & 
\cellcolor{GoodGreen!59.20}0.184 & 
\cellcolor{BadRed!62.90}1.258 \\

L4.weight & 
\cellcolor{GoodGreen!97.55}0.951 & 
\cellcolor{BadRed!30.55}0.611 & 
\cellcolor{GoodGreen!99.95}0.999 & 
\cellcolor{BadRed!7.65}0.153 & 
\cellcolor{GoodGreen!45.85}-0.083 & 
\cellcolor{BadRed!112.40}2.248 \\  

L4.bias & 
\cellcolor{GoodGreen!100.00}1.000 & 
\cellcolor{BadRed!2.60}0.052 & 
\cellcolor{GoodGreen!100.00}1.000 & 
\cellcolor{BadRed!0.30}0.006 & 
\cellcolor{GoodGreen!0.00}-1.000 & 
\cellcolor{BadRed!118.50}2.370 \\  
\bottomrule
\end{tabular}

\vspace{0.2cm}
\parbox{\linewidth}{
\footnotesize
1. \textbf{Color}: \textcolor{GoodGreen}{Green} indicates higher values of CS; \textcolor{BadRed}{Red} represents higher values of ED (color intensity scales with value magnitude) \\
2. \textbf{INI}: Initial parameters of neural networks \\
3. \textbf{Lx}: Neural network layer, where $x$ denotes layer index (0-4)
}
\end{table}

\section{Conclusion}\label{sec:conclusion}

We systematically investigate the influence of Pretrain-DRFWI and Denorm-DRFWI for neural reparameterized FWI methods. Pretrain-DRFWI embeds the initial prior information into the neural networks, while Denorm-DRFWI decouples the initial model from network parameters. Our analysis reveals that Pretrain-DRFWI requires a two-stage complex workflow and is sensitive to negative transfer effects due to mismatches in the objective function. Through experiments, we show that Denorm-DRFWI is more stable and achieves better accuracy in velocity model reconstruction. These findings are further supported by an analysis of the underlying mechanisms.


\ifCLASSOPTIONcaptionsoff
  \newpage
\fi


\end{document}